\documentclass[english]{article}
\usepackage{lmodern}
\usepackage[T1]{fontenc}
\usepackage[latin9]{inputenc}
\usepackage{array}
\usepackage{multirow}
\usepackage{amsbsy}
\usepackage{amstext}
\usepackage{amssymb}
\usepackage{graphicx}

\makeatletter

\providecommand{\tabularnewline}{\\}

\usepackage{microtype}
\usepackage{flushend}
\usepackage{ijcai17}\usepackage{times}\usepackage{helvet}\usepackage{courier}

\usepackage{amsbsy}
\usepackage{algorithmic}
\usepackage[english]{babel}

\makeatother

\usepackage{babel}
\begin{document}

\title{One Size Fits Many: Column Bundle for Multi-X Learning}

\author{Trang Pham, Truyen Tran, Svetha Venkatesh\\
Deakin University, Australia\\
\{\emph{phtra,truyen.tran,svetha.venkatesh}\}\emph{@deakin.edu.au}}

\maketitle
\global\long\def\xb{\boldsymbol{x}}
\global\long\def\yb{\boldsymbol{y}}
\global\long\def\eb{\boldsymbol{e}}
\global\long\def\zb{\boldsymbol{z}}
\global\long\def\hb{\boldsymbol{h}}
\global\long\def\ab{\boldsymbol{a}}
\global\long\def\bb{\boldsymbol{b}}
\global\long\def\cb{\boldsymbol{c}}
\global\long\def\sigmab{\boldsymbol{\sigma}}
\global\long\def\gammab{\boldsymbol{\gamma}}
\global\long\def\alphab{\boldsymbol{\alpha}}
\global\long\def\rb{\boldsymbol{r}}
\global\long\def\fb{\boldsymbol{f}}
\global\long\def\ib{\boldsymbol{i}}
\global\long\def\thetab{\boldsymbol{\theta}}
\global\long\def\pb{\boldsymbol{p}}
\global\long\def\Xbb{\boldsymbol{\text{X}}}
\global\long\def\Ybb{\boldsymbol{\text{Y}}}

\begin{abstract}
Much recent machine learning research has been directed towards leveraging
shared statistics among labels, instances and data views, commonly
referred to as multi-label, multi-instance and multi-view learning.
The underlying premises are that there exist correlations among input
parts and among output targets, and the predictive performance would
increase when the correlations are incorporated. In this paper, we
propose Column Bundle (CLB), a novel deep neural network for capturing
the shared statistics in data. CLB is generic that the same architecture
can be applied for various types of shared statistics by changing
only input and output handling. CLB is capable of scaling to thousands
of input parts and output labels by avoiding explicit modeling of
pairwise relations. We evaluate CLB on different types of data: (a)
multi-label, (b) multi-view, (c) multi-view/multi-label and (d) multi-instance.
CLB demonstrates a comparable and competitive performance in all datasets
against state-of-the-art methods designed specifically for each type.

\end{abstract}

\section{Introduction}

A canonical setting in machine learning is that each training instance
is comprised of an input vector and an output label. However, there
are many learning tasks in which, samples may consist of multiple
input parts and outputs. For example, a video can have multiple types
of input features such as \emph{audio}, \emph{vision} and \emph{text
description}; and in image annotation, an image can be tagged with
multiple concepts such as \emph{horses}, \emph{grass fields} and \emph{tree}s.
For the past decades, machine learning has leveraged shared statistics
between labels, instances, tasks and data views. These have given
rise to fruitful research directions for multi-output data such as
multi-label \cite{elisseeff2001kernel}, multi-task \cite{caruana1997multitask}
and for multi-input data such as multi-view \cite{gonen2011multiple},
multi-instance \cite{dietterich1997solving} and multi-source learning
\cite{crammer2008learning}. Let us simplify the notion by calling
these directions as \emph{multi-X learning}. 

Developed independently is deep learning \textendash{} a scheme emphasizing
learning multiple abstractions \cite{lecun2015deep} through multiple
steps of computation. Central to the recent record-breaking successes
in deep learning is architecture engineering, the art of designing
neural nets that best address the structure of the problems at hand.
There is limited work on neural architectures for leveraging shared
statistics in training data. One example is Column Network (CLN) \cite{pham2016column},
which is a network of thin nets known as columns, designed for leveraging
the relations between data points.

In this paper, we ask a bold question: \emph{Is this possible to build
a generic neural architecture that simultaneously addresses many questions
in multi-X learning}? Inspired by Column Networks, we propose a deep
architecture called \emph{Column Bundle} (CLB), which is a set of
columns connected in a specific way for shared statistics. Unlike
Column Networks where the relations are pre-defined, Column Bundles
integrate separate inputs (e.g., views, parts and instances) indirectly
through a central column. Each input is represented as a mini-column,
which is recurrently connected to the central column. The central
column plays the role of an ``executive function'' in human brain.
In the inference phase, it first serves as a buffer (working memory)
for multiple parts to interact in a sequential manner. The central
column then sends output signals to separate outputs (e.g., label,
task) that are also represented as mini-columns connected to the central
column recurrently. In the learning phase, training signals from (multiple)
labels will propagate back to multiple mini-columns in the input parts.
The mini-columns interact through the central column, therefore the
correlations among inputs or among labels are established through
the bundle. Overall, our system is a single neural architecture that
solves many separately considered sub-problems in recent machine learning.

We evaluate the Column Bundle model on a comprehensive suite of tasks
designed for each of the multi-X problems, as well as data designed
for joint multi-instance/view multi-label learning. Compared with
specific methods designed for each task, our generic architecture
exhibits comparable and competitive results.

Our main contributions are:
\begin{itemize}
\item We suggest the rethinking of recent separate developments in machine
learning, which include multi-X settings, where X is label, task,
view, instance or part. After all, these developments are all based
on the notion that related information should be shared to leverage
the strength of statistics.
\item We design a generic neural architecture called Column Bundle showing
that the same architecture can be used for any multi-X problem by
changing only input and output handling.
\item A comprehensive evaluation, comparing against state-of-the-art algorithms
designed for each setting.
\end{itemize}
The rest of the paper is organized as follows. Relevant work about
shared statistical strength is discussed in Sec.~\ref{sec:Related-work}.
The CLB model is proposed in Sec.~\ref{sec:Method}, followed by
the experimental study (Sec.~\ref{sec:Experiments}). Finally, we
conclude our paper and discuss future work in Sec.~\ref{sec:Discussion}.

\section{Related work \label{sec:Related-work}}

Since this paper touches many fruitful developments of recent machine
learning, we limit ourselves to the most relevant work that aims at
leveraging the shared statistical strength among inputs and outputs
in data.

\emph{Multi-label learning} refers to annotating an instance by more
than one label, usually of the same broad type (e.g., textual tags
for an image) \cite{zhang2014review}. Much work attempted to adapt
single-label algorithms to deal with the multi-label setting, such
as ML-kNN \cite{zhang2007ml} (k-nearest neighbor for multi-label
data) and Rank-SVM \cite{elisseeff2001kernel} (Kernel learning for
multi-label data). Explicit correlations can also be modeled using
conditional random fields as in \cite{ghamrawi2005collective}. Alternatively,
the multi-label learning problem can be transformed into a sequence
of single-label classification problems, where later classifiers in
the sequence take the predictions of previous labels as inputs \cite{cheng2010bayes}.
\emph{Multi-task learning} is similar to multi-label learning, but
more flexible on task definition (e.g., a task per an instance, not
necessarily multiple tasks per an instance) \cite{caruana1997multitask}.
In contrast to having more than one label per data instance, we might
just have one label per several instances, leading to \emph{multi-instance
learning} \cite{dietterich1997solving}. Alternatively, when there
is redundancy in representing the same data, we have a \emph{multi-view}
setting \cite{gonen2011multiple}. This includes \emph{multimodality}
as in multimedia \cite{JMLR:v15:srivastava14b}. One of the earliest
set of algorithms for multi-view learning is co-training, in which
two classifiers for two views learn to agree on the classification
for unlabeled data \cite{kumar2011co}. Alternatively one can resort
to multiple kernel learning in which kernels correspond to different
views \cite{gonen2011multiple}, and subspace learning to discover
a latent subspace shared by multi-views \cite{salzmann2010factorized}.
Occasionally when both multiple outputs and input parts are available,
we have a joint setting, for example, a mixture of multiple instances,
views and labels \cite{feng2017deep}. 

\emph{Neural networks} have been demonstrated to be highly versatile
to handle multiple tasks \cite{collobert2011natural}. Our Column
Bundle bears some similarity to a recent architecture designed for
set-to-set mapping \cite{vinyals2015order}, but we do not sequence
unordered sets and aim at encoding shared information among set elements.
Recently, a deep architecture for multi-instance/multi-label (DeepMIML)
learning has been proposed \cite{feng2017deep}. DeepMIML retains
sub-concepts for each label, and learns a \emph{sub-concept} layer,
which is a 3D tensor to model the matching scores between instances
and sub-concepts of labels. DeepMIML does not model the interactions
among labels or instances. Instead, the sub-concept layer is pooled
twice by the instance dimension and then by the sub-concept dimension
to get a vector of predictions for all labels.

\section{Method \label{sec:Method}}

In this section, we present our main contribution, the Column Bundle
(CLB). Denote by $\text{\textbf{X}}=\left\{ \Xbb^{1},...,\Xbb^{N}\right\} $
the input set of $N$ examples with the corresponding output set $\Ybb=\left\{ \Ybb^{1},...,\Ybb^{N}\right\} $.
Each $\Xbb^{j}$ consists of $M_{V}$ ($M_{V}\geq1$) input parts:
$\Xbb^{j}=\left\{ \xb_{1}^{j}\in\mathbb{R}^{d_{1}},...,\xb_{M_{V}}^{j}\in\mathbb{R}^{d_{M_{V}}}\right\} $
and each $\Ybb^{j}$ consists of $M_{L}$ ($M_{L}\geq1$) outputs:
$\Ybb^{j}=\left\{ y_{1}^{j},...,y_{M_{L}}^{j}\right\} $, where $y_{i}^{j}$
can be either binary or multi-class. We call each input $\xb_{i}^{j}$
or each label $y_{i}^{j}$ as a part. For any unobserved-label example
$\Xbb^{unk}$, a multi-X model predicts $\Ybb^{unk}$ as the set of
outputs for $\Xbb^{unk}$. In what follows, the superscripts indicating
sample indices are removed for clarity.

When $M_{V}>1$ and $M_{L}=1$, the data contain multi-input parts
and a single label, e.g., multi-instance and multi-view learning.
When $M_{V}=1$ and $M_{L}>1$, the data contain only an input part
and multi-outputs, the problem becomes multi-label or multi-task learning.
When both $M_{V}>1$ and $M_{L}>1$, the setting is a combination
of the two previous problems, e.g., multi-label/multi-view and multi-label/multi-instance
learning.

\subsection{Column design \label{subsec:Column-design}}

\begin{figure}[h]
\begin{centering}
\includegraphics[bb=45bp 70bp 752bp 370bp,clip,width=0.95\columnwidth]{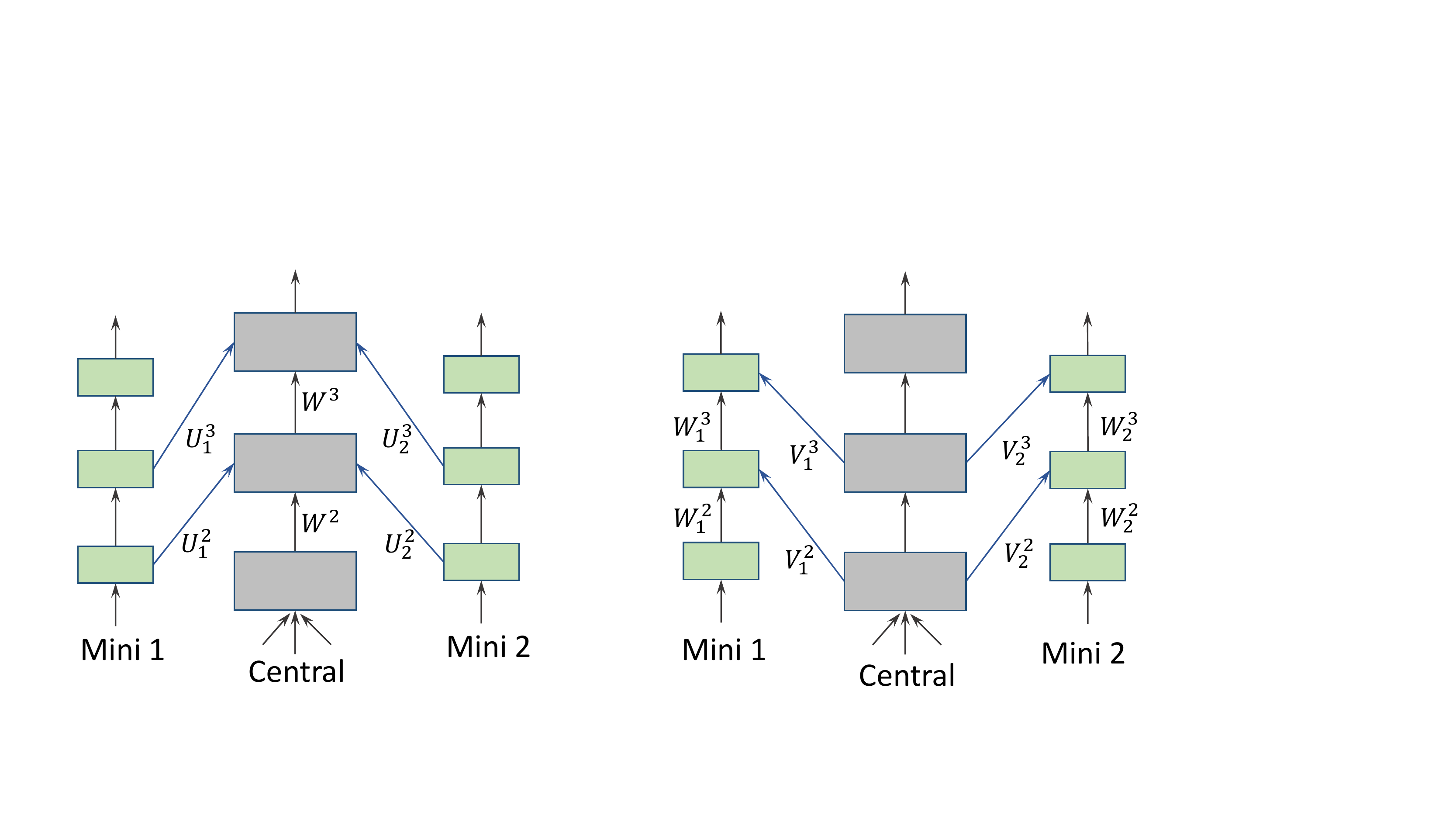}
\par\end{centering}
\caption{A Column Bundle with 2 mini-columns. (Left) Forward flow at the central
column, (Right) Forward flow at the mini-columns. \label{fig:Column-bundle} }
\end{figure}

CLB is inspired by a recent architecture \textendash{} Column Network
\cite{pham2016column}, which is a neural network of thin feedforward
nets known as mini-columns. These columns are inter-connected through
pre-defined relations. We borrow this idea to handle the correlations
among multiple input/output parts in data. However, when the number
of parts grows, modeling the correlation between two parts by a link
is impossible as the number of connections is quadratic in the number
of parts. In a CLB, each part is processed by a mini-column and mini-columns
link to a central column only (See Fig.~\ref{fig:Column-bundle}(Left)).
The central column processes inputs from itself and from the mini-columns,
and then redistributes the output to the mini-columns (See Fig.~\ref{fig:Column-bundle}(Right)).
Through multiple layers, the correlations among mini-columns are established.
With the CLB model, the number of connections is only the number of
parts in data, therefore saving memory and computational cost.

To be more precise, a column bundle consists of a central column and
$M$ mini-columns. Each column is a feedforward neural network of
$T$ hidden layers. Denote by $\hb_{c}^{t}$ the hidden activation
at layer $t$ of the central column and by $\hb_{i}^{t}$ the hidden
activation at layer $t$ of the $i^{th}$ mini-column. At the first
layer in a bundle, each mini-column reads an input part of data and
the central column can read inputs from multiple sources, depending
on the data type (See Sec.~\ref{subsec:CLB-for-data-types} for details)
. At each layer $t$ $(1<t\leq T)$, $\hb_{c}^{t}$ is a non-linear
transformation of its previous hidden state $\hb_{c}^{t-1}$ and previous
hidden states $\hb_{i}^{t-1}$ $(i=1,..,M)$ from all mini-columns.
At each mini-column $i$, $\hb_{i}^{t}$ is a non-linear function
of its previous hidden state $\hb_{i}^{t-1}$ and the previous hidden
state $\hb_{c}^{t-1}$ from the central column. The transformations
are written as follows

\begin{eqnarray}
\hb_{c}^{t} & = & g\left(W^{t}\hb_{c}^{t-1}+\frac{1}{M}\sum_{i=1}^{M}U_{i}^{t}\hb_{i}^{t-1}\right)\label{eq:central_hidden}\\
\hb_{i}^{t} & = & g\left(W_{i}^{t}\hb_{i}^{t-1}+V_{i}^{t}\hb_{c}^{t-1}\right)\label{eq:mini_hidden}
\end{eqnarray}
where $W^{t},U_{i}^{t},W_{i}^{t}\text{ and }V_{i}^{t}$ are weight
matrices at layer $t$. Here, biases are omitted for clarity. At the
top layer, the hidden state $\hb_{c}^{T}$ can be returned for further
purposes.

\subsubsection*{Highway Networks as columns}

Each column (mini- or central one) in a bundle can be any feedforward
neural network. However, training traditional deep feedforward neural
nets is difficult as non-linear functions prevent data signals and
gradients from passing easily through the network. In our implementation,
we opt for Highway Networks \cite{srivastava2015training}, a solution
addressing the problem by adding gates that let previous states propagate
partly through layers:

\[
\hb^{t}=\alphab_{1}^{t}*\tilde{\hb}^{t}+\alphab_{2}^{t}*\hb^{t-1}
\]
where $\tilde{\hb}^{t}$ is a nonlinear candidate function of inputs
at layer $t$ and $\alphab_{1}^{t},\alphab_{2}^{t}\in(\boldsymbol{0},\boldsymbol{1})$
are learnable gates. This enables input signals and error gradients
to propagate through very deep networks. We set the gate $\alphab_{1}^{t}$
as a sigmoid function and the gate $\alphab_{2}^{t}=1-\alphab_{1}^{t}$
following \cite{srivastava2015training}.

In a CLB, the candidate functions $\tilde{\hb}_{c}^{t}$ for the central
column $c$ and $\tilde{\hb}_{i}^{t}$ for each mini-column $i$ are
computed using Eq.~(\ref{eq:central_hidden}) and Eq.~(\ref{eq:mini_hidden}),
respectively. The gates $\alphab_{c1}^{t}$ at the central column
and $\alphab_{i1}^{t}$ at each mini-column are modeled as:

\begin{eqnarray}
\alphab_{c1}^{t} & = & \sigma\left(W_{\alphab}^{t}*\hb_{c}^{t-1}+\frac{1}{M}\sum_{i=1}^{M}U_{\alphab i}^{t}\hb_{i}^{t-1}\right)\label{eq:central_gate}\\
\alphab_{i1}^{t} & = & \sigma\left(W_{\alphab i}^{t}\hb_{i}^{t-1}+V_{\alphab i}^{t}\hb_{c}^{t-1}\right)\label{eq:mini_gate}
\end{eqnarray}

\subsubsection*{Parameter sharing among layers}

The number of parameters grows with the number of hidden layers in
feedforward networks and with the number of labels/inputs in data.
This may cause overfitting in very deep networks. To address this,
work has been done using the idea of parameter sharing in Recurrent
Neural Networks (RNNs) \cite{pham2016faster,pham2016column,liao2016bridging},
that is hidden layers share the same set of parameters. This helps
the depth of neural nets to grow whilst saving parameters. When parameters
are shared among layers, the weights in Eq.~(\ref{eq:central_hidden},~\ref{eq:mini_hidden})
become: $W^{1}=...=W^{T}=W$, $U_{i}^{1}=...=U_{i}^{T}=U_{i}$, $W_{i}^{1}=...=W_{i}^{T}=W_{i}$
and $V_{i}^{1}=...=V_{i}^{T}=V_{i}$ ($i=1...M$). It is similar for
the weight matrices of the gates in Eqs.~(\ref{eq:central_gate},~\ref{eq:mini_gate}).

\subsection{CLB for different multi-X settings \label{subsec:CLB-for-data-types}}

There are different types of multi-X data: \emph{multi-input} (e.g.,
multi-instance, multi-view), \emph{multi-output} (e.g., multi-task,
multi-label) and \emph{a combination of both} (e.g., multi-instance/multi-label,
multi-view/multi-label). Our CLB model can work on all of these types
without changing the structure. For each type of data, we only need
to set up the inputs and the outputs for the CLB. In this subsection,
we describe the ways that inputs and outputs are handled for each
multi-X problem.

\begin{figure}[h]
\centering{}\includegraphics[bb=150bp 40bp 670bp 490bp,clip,width=0.97\columnwidth]{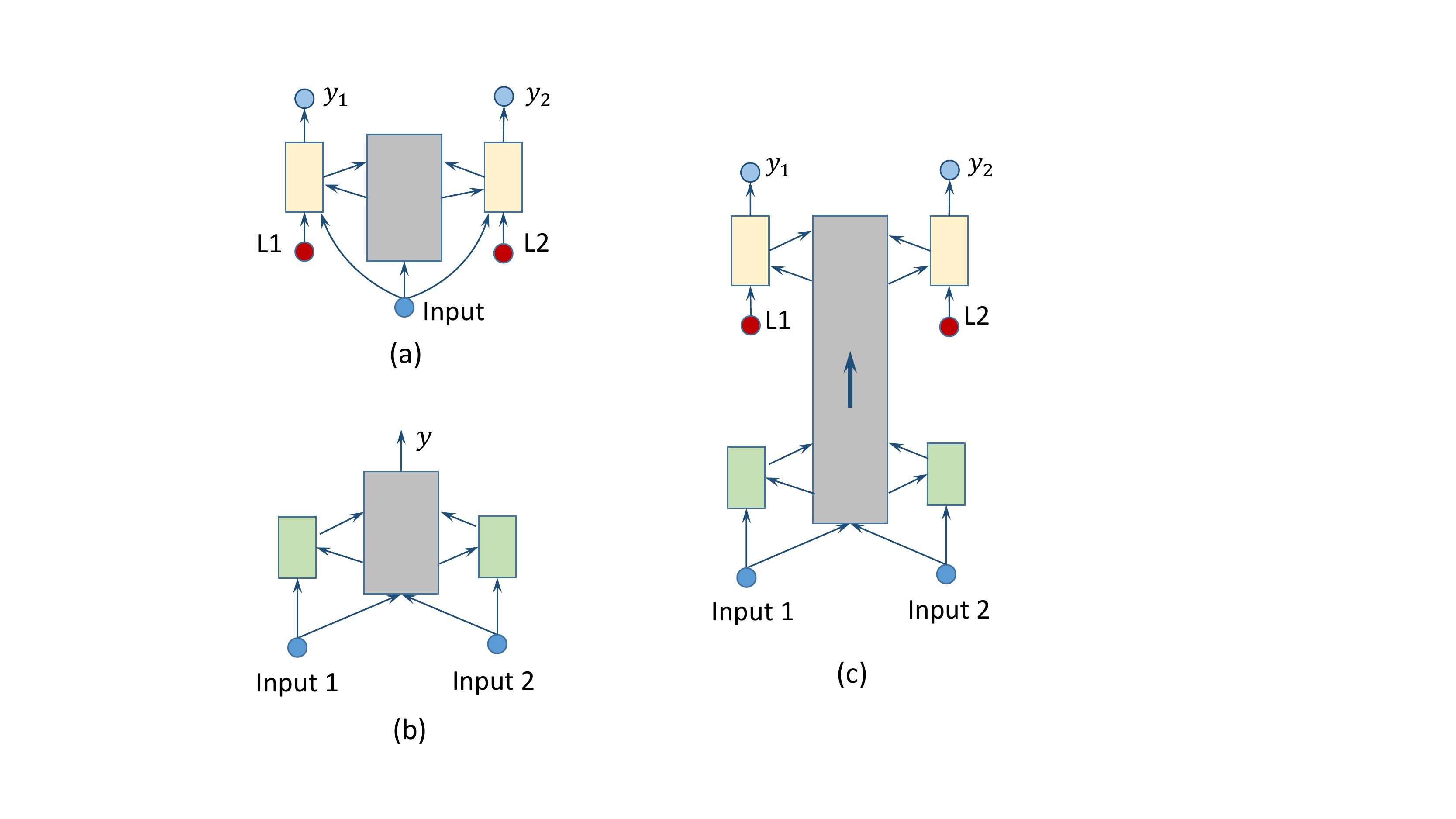}\caption{CLB models for multi-X settings. (a) Multi-outputs (b) Multi-inputs
(c) Multi-inputs/Multi-outputs \label{fig:CLB-multi-type}}
\end{figure}

\subsubsection*{Multi-outputs }

The CLB for multi-output problems (i.e., multi-task, multi-label learning)
is shown in Fig.~\ref{fig:CLB-multi-type}a. The central column processes
and distributes the input signals to the mini-columns. The mini-columns
process these signals and send the output signals back to the central
column. The process repeats recurrently along the layers. At the top,
each mini-column predicts its own output. Let $M_{L}$ be the number
of labels, $y_{i}$ be the class of the label $i^{th}$ and $\xb$
be the feature vector for a sample. For simplicity, at the first layer,
the central column and all the mini-columns read the feature vector
\begin{eqnarray}
\hb_{c}^{1} & = & g\left(W^{1}\xb\right)\label{eq:ML-first-central-hidden}\\
\hb_{i}^{1} & = & g\left(V_{i}^{1}\xb\right)\label{eq:ML-first-mini-hidden}
\end{eqnarray}
for $i=1...M_{L}$ and process the data signals as described in Sec.~\ref{subsec:Column-design}.
The mini-column $i^{th}$ predicts a class for the label $i^{th}$
($i=1...M_{L}$) as normal: $P_{i}\left(y_{i}=1\right)=\sigma\left(Z\hb_{i}^{T}\right)$
for binary classification and $P_{i}\left(y_{i}=c\right)=\text{softmax}\left(Q\hb_{i}^{T}\right)$
for multi-class classification. The loss function for a sample is
the sum of the negative-log likelihood of all labels

\begin{equation}
L=-\sum_{i=1}^{M_{L}}\text{log}\left(P_{i}\left(y_{i}\right)\right)\label{eq:ML-loss-func}
\end{equation}

\subsubsection*{Multi-inputs}

The model for multi-input data (e.g., multi-instance, multi-view)
is illustrated in Fig.~\ref{fig:CLB-multi-type}b. The central column
recurrently receives input signals from the mini-columns. At the top
layer, the central column predicts the output of the sample. Let $M_{V}$
be the number of input parts and $\xb_{1},\xb_{2},..,\xb_{M_{V}}$
be feature vectors of these inputs. At the first layer of CLB, each
mini-column reads a feature vector of an input part and the central
column reads all inputs. 
\begin{eqnarray}
\hb_{c}^{1} & = & g\left(\frac{1}{M_{V}}\sum_{i=1}^{M}U_{i}^{1}\xb_{i}\right)\label{eq:MV-first-central-hidden}\\
\hb_{i}^{1} & = & g\left(W_{i}^{t}\xb_{i}\right)\label{eq:MV-first-mini-hidden}
\end{eqnarray}
for $i=1...M_{V}$. The hidden activation at the top layer of the
central column $\hb_{c}^{T}$ can be used for predicting the output.

\subsubsection*{Multi-inputs/multi-outputs}

This setting occurs when both multi-inputs and multi-outputs are available
in data, for example, multi-instance/multi-label and multi-view/multi-label
data. The model for multi-input/multi-output setting is simply a combination
of the two settings described above (See Fig.~\ref{fig:CLB-multi-type}c
for illustration). The hidden activation at the top layer of the central
column for multi-inputs is passed to the central column for multi-outputs. 

\subsection{Scaling up for thousands of parts \label{subsec:Scaling-up}}

Each mini-column processes an input or output in data. These inputs/outputs
may be of different types, thus normally each mini-column has it own
parameters, except multi-instance learning where instances are of
the same type. However, some datasets contain many input parts or
labels, for example, MediaMill dataset (See Sec.~\ref{subsec:Multi-label})
with 101 labels. The number of parameters for the mini-columns in
these cases would be extremely huge. To tackle this issue, mini-columns
can share parameters (it is different from parameter sharing among
layers described in Sec.~\ref{subsec:Column-design}). That means
the weight matrices in Eqs.~(\ref{eq:central_hidden},~\ref{eq:mini_hidden})
become: $U_{1}=...=U_{M}$, $W_{1}=...=W_{M}$ and $V_{1}=...=V_{M}$.
It is similar for the weight matrices of the gates in Eq.~(\ref{eq:central_gate},~\ref{eq:mini_gate}).

For \textbf{\emph{multi-label}} problems, if the parameters of the
mini-columns are shared, the mini-columns return the same output as
they receive the same data signals from the central column and process
the information by the same weights. To resolve the issue, we borrow
the idea of label embedding, which is used to address the problem
of multi-class learning with many classes \cite{amit2007uncovering,bengio2010label}.
In our work, instead of learning an embedding matrix for classes,
the CLB model retains embedded vectors for label indices and then
feeds these vectors to the mini-columns as inputs. The model learns
an embedding matrix $E\in\mathbb{R}^{d_{e}\times M_{L}}$, where $M_{L}$
is the number of labels, $d_{e}$ is the embedding dimension and the
column $E_{i}$ is the embedded vector of the label $i^{th}$. Each
mini-column $i$ reads the vector $E_{i}$ as the input signal for
the label $i^{th}$. The Eq.~(\ref{eq:ML-first-central-hidden},~\ref{eq:ML-first-mini-hidden})
are now replaced by:
\begin{eqnarray}
\hb_{c}^{1} & = & g\left(W^{1}\xb+\frac{1}{M_{L}}\sum_{i=1}^{M_{L}}U_{i}^{1}E_{i}\right)\label{eq:ML-first-central-hidden-embed}\\
\hb_{i}^{1} & = & g\left(W_{i}^{1}E_{i}+V_{i}^{1}\xb\right)\label{eq:ML-first-mini-hidden-embed}
\end{eqnarray}

For\emph{ }\textbf{\emph{multi-input}} problems, input parts are feature
vectors in different dimensions. Before being passed to the mini-columns,
input parts are projected into the same vector space.

\section{Experiments \label{sec:Experiments}}

We demonstrate the effectiveness of CLB in handling various types
of multi-X settings by experimenting on Multi-label, Multi-view, Multi-view/Multi-label
and Multi-instance datasets.

\subsection{Implementation}

For all experiments with CLB, dropout is applied before and after
the recurrent layers of the model. To handle data imbalance, each
class is weighted by the log of the multiplicative inverse of its
frequency. This implies that the model attends more to samples from
an under-represented class. Each dataset is divided into 3 separated
sets: training, validation and test sets. The learning rate starts
at 0.001. After 10 epochs, the learning rate is divided by 2 if the
model cannot find a better result on the validation set. Learning
is terminated after 4 times of halving the learning rate or after
500 epochs. For hyper-parameter tuning, we set the number of hidden
layers by 10 and search for (i) hidden dimensions of the central column,
(ii) hidden dimensions of the mini-columns (we set all the mini-columns
in the same dimension) , (iii) embedding dimension for label embedding
and (iv) optimizers: Adam or RMSprop. The best setting is chosen by
the validation set and the results of the test set are reported as
the mean result of 5 runs. Code for CLB model can be found on Github\footnote{link omitted for review}.

\subsection{Multi-label \label{subsec:Multi-label}}

\subsubsection*{Datasets}

Our first set of experiments test the CLB as a model for multi-label
learning (See Sec.~\ref{subsec:CLB-for-data-types} and Fig.~\ref{fig:CLB-multi-type}a).
We use 3 datasets with different number of labels and density. The
first dataset is Movielens Latest Dataset\footnote{https://grouplens.org/datasets/movielens/}.
The task is to predict genres for each movie given plot summary. After
removing all movies without plot summary, the dataset contains 18,352
movies. The other two datasets - tmc2007 and MediaMill - are downloaded
from Mulan website\footnote{http://mulan.sourceforge.net/datasets-mlc.html}.
The statistics of the three datasets are reported in Table~\ref{tab:Multi-label-datasets}.

\begin{table}[h]
\centering{}%
\begin{tabular}{|l|c|r|c|c|}
\hline 
Dataset & n\_samples & n\_feats & n\_labels & Density\tabularnewline
\hline 
\hline 
Movielens & 18,352 & 1000 & 9 & 0.212\tabularnewline
tmc2007 & 28,596 & 500 & 22 & 0.098\tabularnewline
MediaMill & 43,907 & 120 & 101 & 0.046\tabularnewline
\hline 
\end{tabular}\caption{Statistics of Multi-label datasets\label{tab:Multi-label-datasets}}
\end{table}

\subsubsection*{Baselines and experiment settings}

For comparison, we employed baseline methods specifically designed
for (a) multi-label learning and (b) deep neural nets for training
each label separately. For the former, we used 3 methods listed below:
\begin{itemize}
\item Probabilistic Classifier Chains (PCC) \cite{cheng2010bayes}
\item Learning label-specific features for multi-label classification (LLSF)
\cite{huang2015learning}
\item Back Propagation Neural Network for multi-label (BPNN) - a method
in Meka toolkit \cite{read2016meka}
\end{itemize}
For deep neural nets, we implemented Highway Network with shared parameters
among layers (HWN), trained each label separately with a neural network,
and then combined the results of all the labels.

In the CLB model, the parameters of the mini-columns are not shared
for datasets with small number of labels (Movielens dataset), and
shared using label embedding (See Sec.~\ref{subsec:Scaling-up})
for datasets with large number of labels (tmc2007 and MediaMill datasets).

\subsubsection*{Results}

\begin{table*}
\centering{}%
\begin{tabular}{|l|cc|cc|cc|}
\hline 
\multirow{2}{*}{Method} & \multicolumn{2}{c|}{Movielens} & \multicolumn{2}{c|}{tmc2007} & \multicolumn{2}{c|}{MediaMill}\tabularnewline
\cline{2-7} 
 & MicroF1 & HLoss & MicroF1 & HLoss & MicroF1 & HLoss\tabularnewline
\hline 
\hline 
PCC & \textbf{55.6} & 0.229 & 73.2 & 0.058 & \emph{56.0} & 0.035\tabularnewline
LLSF & 51.8 & 0.208 & 64.9 & 0.064 & 54.0 & \textbf{0.031}\tabularnewline
BPNN & 53.8 & 0.196 & 66.9 & 0.067 & 55.4 & 0.039\tabularnewline
HWN & 53.0 & \textbf{0.190} & \emph{76.0} & \emph{0.053} & 22.4 & 0.035\tabularnewline
\hline 
CLB & \emph{54.3} & \emph{0.191} & \textbf{76.5} & \textbf{0.049} & \textbf{56.7} & \emph{0.032}\tabularnewline
\hline 
\end{tabular}\caption{Performance on multi-label datasets, reported in Micro F1-score (\%,
the higher the better) and Hamming Loss (HLoss, the smaller the better).
The best score is in \textbf{bold} and the second best is in \emph{italic}.\label{tab:Results-multi-labels}}
\end{table*}

Table~\ref{tab:Results-multi-labels} summarizes the Micro F1-scores
and Hamming Loss obtained by all models on the three datasets. CLB
is stable, comparable or better than other baselines in all three
datasets. On Movielens, although PCC outperforms CLB on Micro F1-score
and HWN outperforms CLB on Hamming Loss, the two baselines fail to
compete with our method on the other metric. On tmc2007 dataset, CLB
achieves a F1-score of 76.5\%, followed by 76\% and 73.2\%, achieved
by HWN and PCC. HWN performs quite well with small number of labels
and high density but it fails to handle MedialMill dataset with low
label density and 101 labels. 

\begin{figure}[h]
\centering{}\includegraphics[bb=58bp 35bp 700bp 315bp,clip,width=0.93\columnwidth]{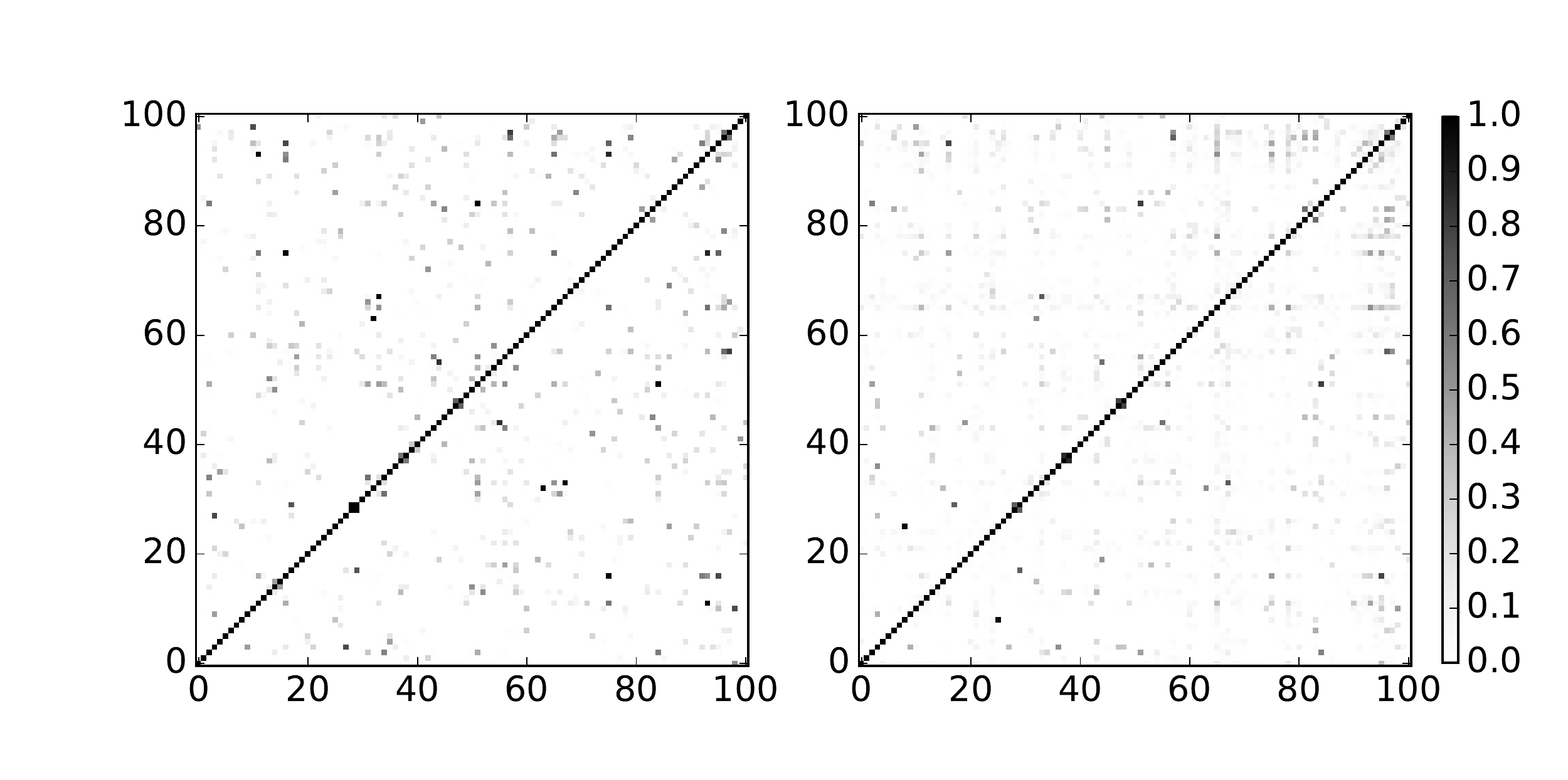}\caption{(Left) Pairwise cosine similarity matrix of label embedded vectors,
(Right) Pairwise correlation matrix of 101 labels for MediaMill dataset.
Values are normalized so both matrices are in the same range {[}0,
1{]}.\label{fig:correlation-mediamill}}
\end{figure}

Fig.~\ref{fig:correlation-mediamill}(Left) shows the pairwise cosine
similarity of label embedded vectors and (Right) The pairwise correlations
of 101 labels for MediaMill dataset. The correlations are computed
based on label co-occurrence. The two matrices are quite similar.
It suggests that the learned label embedded vectors somehow capture
the correlations among labels.

\subsection{Multi-view and Multi-view/Multi-label}

\subsubsection*{Dataset}

For multi-view learning (See Sec.~\ref{subsec:CLB-for-data-types}
and Fig.~\ref{fig:CLB-multi-type}b), we evaluate our model on Youtube
dataset \cite{madani2013using}. The dataset consists of features
and labels of 120,000 videos. The task is to classify each video into
one of 31 classes where each class (except class 31) corresponds to
a video game and class 31 corresponds to other games that do not belong
to any of the 30 games. Each video sample is described by 13 views,
from 3 feature families: textual, visual and auditory features. In
our experiments, for fair comparison with other baselines, we use
textual (3 views) and visual (5 views) features only. Text features
are in bag-of-word representation. We preprocessed the data by removing
all words with doc-frequency smaller than 0.01.

For multi-view/multi-label learning (See Sec.~\ref{subsec:CLB-for-data-types}
and Fig.~\ref{fig:CLB-multi-type}c), we use NUS-WIDE dataset \cite{chua2009nus}.
The dataset contains 269,648 images associated with tags from Flickr
and six types of low-level features (visual and textual) extracted
from these images. Visual features include 64-D color histogram, 144-D
color correlogram, 73-D edge direction histogram, 128-D wavelet texture
and 225-D block-wise color moments extracted over 5x5 fixed grid partitions,
and textual features are 500-D bag-of-word descriptions. We only use
these low-level features for experiments. Each image can be labeled
with some of 81 concepts (e.g., \emph{water}, \emph{buildings},\emph{
mountain}, \emph{cars}, etc.). The label density is only 0.023.

\subsubsection*{Baselines and experiment settings}

We compare CLB against two baselines: (i) a highway net that reads
the concatenation of all views as input (HWN) and (ii) deep Boltzmann
Machines for Multimodal learning (BMM) \cite{JMLR:v15:srivastava14b}.
For the latter, we use the code provided by the authors in which all
textual feature vectors are concatenated into a view and all visual
feature vectors are concatenated into another view. BMM is used to
extract a unified representation for the two views and this representation
is then fed to a Highway Network as input features for classification.
We set the dimension of extracted feature vectors to 1024 and tune
the hidden layer size of the Highway Network.

For a fair comparison with BMM, we evaluate the performance of CLB
in 2 settings of data: (i) 2-view setting where views in the same
types are concatenated, same as in BMM method and (ii) all-view setting
where each mini-column processes a view.

For NUS-WIDE dataset, we use the CLB model as in Fig.~\ref{fig:CLB-multi-type}c.
The two baselines train each label separately and then combine the
results of all labels.

\subsubsection*{Results}

The performance of CLB and the baselines on Youtube and NUS-WIDE datasets
is reported in Table~\ref{tab:Results-multi-views}. Both settings
of CLB beat all baselines. The CLB model performs on the all-view
setting slightly better than it does on the 2-view setting. 

\begin{figure}[h]
\centering{}%
\begin{tabular}{c}
\includegraphics[bb=45bp 75bp 670bp 285bp,clip,width=0.9\columnwidth]{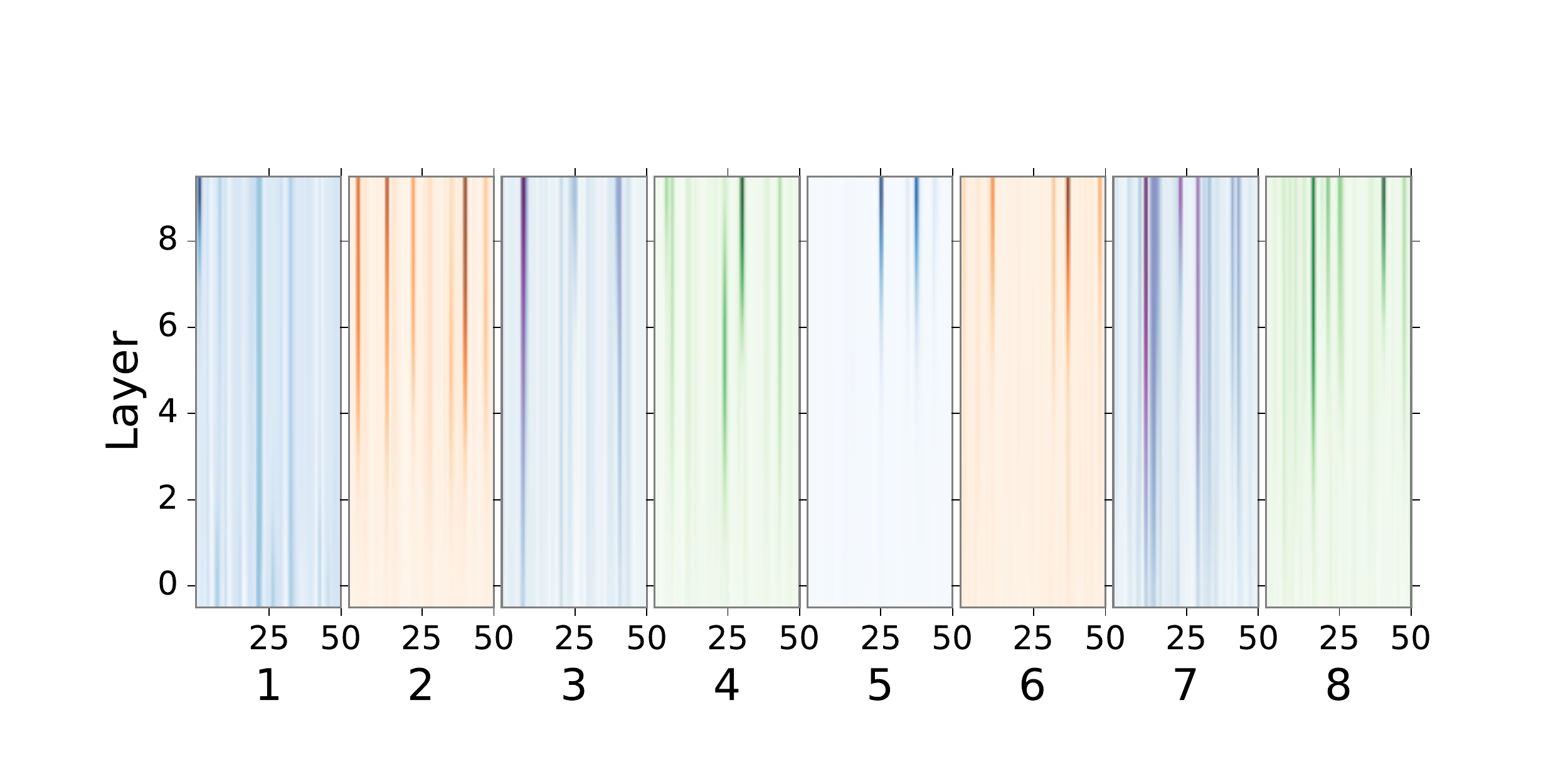}\tabularnewline
\includegraphics[bb=45bp 35bp 670bp 285bp,clip,width=0.9\columnwidth]{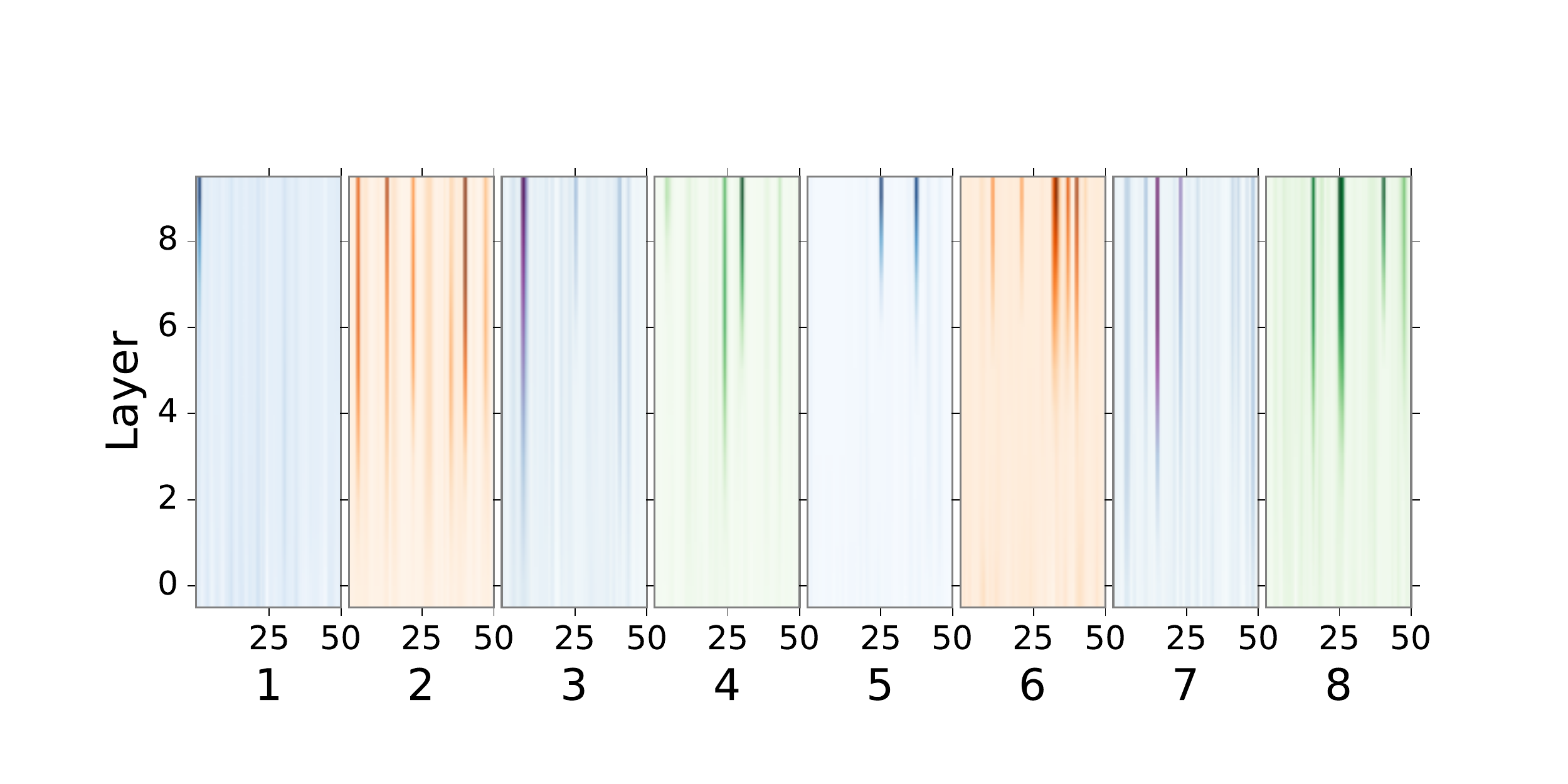}\tabularnewline
\end{tabular}\caption{The dynamics of 50 units through 10 hidden layers of 8 mini columns
of two random samples in the same category for Youtube dataset. Darker
areas for larger values. (Best view in color)\label{fig:youtube_hidden}}
\end{figure}

In Youtube dataset, we randomly choose 2 samples in the same class
and visualize their hidden states through 10 layers of 8 mini-columns
(See Fig.~\ref{fig:youtube_hidden}). It is interesting that each
pair of mini-columns from the two samples are in the similar patterns.

\begin{table}[h]
\centering{}%
\begin{tabular}{|l|cc|cc|}
\hline 
\multirow{2}{*}{Method} & \multicolumn{2}{c|}{Youtube} & \multicolumn{2}{c|}{NUS-WIDE}\tabularnewline
\cline{2-5} 
 & MicroF1 & HLoss & MicroF1 & HLoss\tabularnewline
\hline 
\hline 
HWN & 97.3  & 0.027 & 53.1 & 0.022\tabularnewline
2views-BMM & 95.2 & 0.048 & 50.0 & 0.023\tabularnewline
\hline 
2views-CLB & 97.9 & 0.021 & 56.9 & \textbf{0.019}\tabularnewline
CLB & \textbf{98.0} & \textbf{0.020} & \textbf{57.7} & \textbf{0.019}\tabularnewline
\hline 
\end{tabular}\caption{Results on Youtube and NUS-WIDE datasets, reported in Micro F1-score
(\%) and Hamming Loss (HLoss). \label{tab:Results-multi-views}}
\end{table}

\subsection{Multi-instance}

We evaluate CLB on IMDB review sentiment classification\footnote{http://ai.stanford.edu/\textasciitilde{}amaas/data/sentiment/}
in a multi-instance setting. The dataset contains 25,000 reviews for
training, 25,000 reviews for testing and additional unlabeled data.
Each review is considered as a bag of sentences. Representation of
sentences is learned from both unlabeled and labeled reviews using
\emph{paragraph2vec} \cite{le2014distributed} on the \emph{gensim}\footnote{https://radimrehurek.com/gensim/}
toolkit. After this step, each sentence is represented as a feature
vector of 100 units. 

For comparison, 3 baselines for multi-instance learning are employed:
the two first algorithms are \textbf{miVLAD} and \textbf{miFV} \cite{wei2014scalable},
and the third is \textbf{MI-Net} \cite{wang2016revisiting} using
highway network. In a MI-Net, instances of each bag are passed to
a highway network and hidden states of instances at the top layer
are pooled (mean and max pooling) to a vector, which is then used
for label prediction. Max pooling works badly on this dataset, therefore,
only the results of MI-Net with mean pooling are reported. Table~\ref{tab:Results-multi-instance}
summaries the results.

\begin{table}[h]
\centering{}%
\begin{tabular}{|l|cc|}
\hline 
\multirow{2}{*}{Method} & \multicolumn{2}{c|}{IMDB}\tabularnewline
\cline{2-3} 
 & F1 & HLoss\tabularnewline
\hline 
\hline 
miVLAD & 81.6 & 0.186\tabularnewline
miFV & 83.9 & 0.162\tabularnewline
MI-Net & 84.7 & 0.158\tabularnewline
\hline 
CLB & \textbf{85.4} & \textbf{0.150}\tabularnewline
\hline 
\end{tabular}\caption{Performance of CLB and the baseline on IMDB sentiment classification
in multi-instance setting. \label{tab:Results-multi-instance}}
\end{table}

\section{Conclusion \label{sec:Discussion}}

We have proposed \emph{Column Bundle} (CLB), a deep neural network
model for leveraging shared statistics in multi-X problems, where
\emph{X} stands for \emph{view}, \emph{label} and \emph{instance}
and \emph{any combination} of these. This is the capability not seen
in existing work. A CLB consists of a bundle of mini-columns connected
to a central column. Each mini-column represents an input part or
an output target, and the central column acts as a hub for the mini-columns
to exchange information. The structure effectively captures the correlations
among the mini-columns without explicitly defining pairwise links.
With sharing parameters among the mini-columns, CLB is capable of
modeling data with a huge number of inputs and labels, e.g., image
tagging in vision with \emph{thousands} of tags. CLB is efficient
as it has only linear complexity in the number of inputs/outputs in
both training and inference. Empirically, CLB demonstrates a competitive
and comparable performance against rivals designed specifically for
multi-label, multi-view, multi-instance learning and multi-view/multi-label
learning. 

CLB opens rooms for future work. For example, a database record of
multiple fields can be represented naturally using a CLB, hence we
can carry out common tasks such as retrieval, record linkage or database
completion. One might also learn an attention mechanism so that the
model can assign more weight to important parts. We can also exploit
variable computation steps in mini-columns so that some easy and linear
tasks (inputs) only need short decoding (encoding).

\section{Acknowledgement}

This work is partially supported by the Telstra-Deakin Centre of Excellence
in Big Data and Machine Learning

\bibliographystyle{named}
\bibliography{../bibs/ME,../bibs/truyen,../bibs/trang}

\begin{thebibliography}{}

\bibitem[\protect\citeauthoryear{Amit \bgroup \em et al.\egroup
  }{2007}]{amit2007uncovering}
Yonatan Amit, Michael Fink, Nathan Srebro, and Shimon Ullman.
\newblock Uncovering shared structures in multiclass classification.
\newblock In {\em Proceedings of the 24th international conference on Machine
  learning}, pages 17--24. ACM, 2007.

\bibitem[\protect\citeauthoryear{Bengio \bgroup \em et al.\egroup
  }{2010}]{bengio2010label}
Samy Bengio, Jason Weston, and David Grangier.
\newblock Label embedding trees for large multi-class tasks.
\newblock In {\em Advances in Neural Information Processing Systems}, pages
  163--171, 2010.

\bibitem[\protect\citeauthoryear{Caruana}{1997}]{caruana1997multitask}
R.~Caruana.
\newblock {Multitask learning}.
\newblock {\em Machine Learning}, 28(1):41--75, 1997.

\bibitem[\protect\citeauthoryear{Cheng \bgroup \em et al.\egroup
  }{2010}]{cheng2010bayes}
Weiwei Cheng, Eyke H{\"u}llermeier, and Krzysztof~J Dembczynski.
\newblock Bayes optimal multilabel classification via probabilistic classifier
  chains.
\newblock In {\em Proceedings of the 27th international conference on machine
  learning (ICML-10)}, pages 279--286, 2010.

\bibitem[\protect\citeauthoryear{Chua \bgroup \em et al.\egroup
  }{2009}]{chua2009nus}
T.S. Chua, J.~Tang, R.~Hong, H.~Li, Z.~Luo, and Y.~Zheng.
\newblock {NUS-WIDE: A real-world web image database from National University
  of Singapore}.
\newblock In {\em Proceedings of the ACM International Conference on Image and
  Video Retrieval}, page~48. ACM, 2009.

\bibitem[\protect\citeauthoryear{Collobert \bgroup \em et al.\egroup
  }{2011}]{collobert2011natural}
Ronan Collobert, Jason Weston, L{\'e}on Bottou, Michael Karlen, Koray
  Kavukcuoglu, and Pavel Kuksa.
\newblock Natural language processing (almost) from scratch.
\newblock {\em The Journal of Machine Learning Research}, 12:2493--2537, 2011.

\bibitem[\protect\citeauthoryear{Crammer \bgroup \em et al.\egroup
  }{2008}]{crammer2008learning}
K.~Crammer, M.~Kearns, and J.~Wortman.
\newblock {Learning from multiple sources}.
\newblock {\em The Journal of Machine Learning Research}, 9:1757--1774, 2008.

\bibitem[\protect\citeauthoryear{Dietterich \bgroup \em et al.\egroup
  }{1997}]{dietterich1997solving}
T.G. Dietterich, R.H. Lathrop, and T.~Lozano-P{\'e}rez.
\newblock Solving the multiple instance problem with axis-parallel rectangles.
\newblock {\em Artificial Intelligence}, 89(1-2):31--71, 1997.

\bibitem[\protect\citeauthoryear{Elisseeff and
  Weston}{2001}]{elisseeff2001kernel}
A.~Elisseeff and J.~Weston.
\newblock A kernel method for multi-labelled classification.
\newblock {\em Advances in neural information processing systems}, 14:681--687,
  2001.

\bibitem[\protect\citeauthoryear{Feng and Zhou}{2017}]{feng2017deep}
Ji~Feng and Zhi-Hua Zhou.
\newblock Deep miml network.
\newblock {\em AAAI}, 2017.

\bibitem[\protect\citeauthoryear{Ghamrawi and
  McCallum}{2005}]{ghamrawi2005collective}
N.~Ghamrawi and A.~McCallum.
\newblock {Collective multi-label classification}.
\newblock In {\em Proceedings of the 14th ACM international conference on
  Information and knowledge management}, pages 195--200. ACM New York, NY, USA,
  2005.

\bibitem[\protect\citeauthoryear{G{\"o}nen and
  Alpayd{\i}n}{2011}]{gonen2011multiple}
Mehmet G{\"o}nen and Ethem Alpayd{\i}n.
\newblock Multiple kernel learning algorithms.
\newblock {\em Journal of Machine Learning Research}, 12(Jul):2211--2268, 2011.

\bibitem[\protect\citeauthoryear{Huang \bgroup \em et al.\egroup
  }{2015}]{huang2015learning}
Jun Huang, Guorong Li, Qingming Huang, and Xindong Wu.
\newblock Learning label specific features for multi-label classification.
\newblock In {\em Data Mining (ICDM), 2015 IEEE International Conference on},
  pages 181--190. IEEE, 2015.

\bibitem[\protect\citeauthoryear{Kumar and Daum{\'e}}{2011}]{kumar2011co}
Abhishek Kumar and Hal Daum{\'e}.
\newblock A co-training approach for multi-view spectral clustering.
\newblock In {\em Proceedings of the 28th International Conference on Machine
  Learning (ICML-11)}, pages 393--400, 2011.

\bibitem[\protect\citeauthoryear{Le and Mikolov}{2014}]{le2014distributed}
Quoc~V Le and Tomas Mikolov.
\newblock Distributed representations of sentences and documents.
\newblock {\em ICML}, 2014.

\bibitem[\protect\citeauthoryear{LeCun \bgroup \em et al.\egroup
  }{2015}]{lecun2015deep}
Yann LeCun, Yoshua Bengio, and Geoffrey Hinton.
\newblock Deep learning.
\newblock {\em Nature}, 521(7553):436--444, 2015.

\bibitem[\protect\citeauthoryear{Liao and Poggio}{2016}]{liao2016bridging}
Qianli Liao and Tomaso Poggio.
\newblock Bridging the gaps between residual learning, recurrent neural
  networks and visual cortex.
\newblock {\em arXiv preprint arXiv:1604.03640}, 2016.

\bibitem[\protect\citeauthoryear{Madani \bgroup \em et al.\egroup
  }{2013}]{madani2013using}
Omid Madani, Manfred Georg, and David~A Ross.
\newblock On using nearly-independent feature families for high precision and
  confidence.
\newblock {\em Machine learning}, 92(2-3):457--477, 2013.

\bibitem[\protect\citeauthoryear{Pham \bgroup \em et al.\egroup
  }{2016}]{pham2016faster}
Trang Pham, Truyen Tran, Dinh Phung, and Svetha Venkatesh.
\newblock Faster training of very deep networks via p-norm gates.
\newblock {\em ICPR}, 2016.

\bibitem[\protect\citeauthoryear{Pham \bgroup \em et al.\egroup
  }{2017}]{pham2016column}
Trang Pham, Truyen Tran, Dinh Phung, and Svetha Venkatesh.
\newblock Column networks for collective classification.
\newblock {\em AAAI}, 2017.

\bibitem[\protect\citeauthoryear{Read \bgroup \em et al.\egroup
  }{2016}]{read2016meka}
Jesse Read, Peter Reutemann, Bernhard Pfahringer, and Geoff Holmes.
\newblock Meka: a multi-label/multi-target extension to weka.
\newblock {\em Journal of Machine Learning Research}, 17(21):1--5, 2016.

\bibitem[\protect\citeauthoryear{Salzmann \bgroup \em et al.\egroup
  }{2010}]{salzmann2010factorized}
Mathieu Salzmann, Carl~Henrik Ek, Raquel Urtasun, and Trevor Darrell.
\newblock Factorized orthogonal latent spaces.
\newblock In {\em AISTATS}, pages 701--708, 2010.

\bibitem[\protect\citeauthoryear{Srivastava and
  Salakhutdinov}{2014}]{JMLR:v15:srivastava14b}
Nitish Srivastava and Ruslan Salakhutdinov.
\newblock Multimodal learning with deep boltzmann machines.
\newblock {\em Journal of Machine Learning Research}, 15:2949--2980, 2014.

\bibitem[\protect\citeauthoryear{Srivastava \bgroup \em et al.\egroup
  }{2015}]{srivastava2015training}
Rupesh~K Srivastava, Klaus Greff, and J{\"u}rgen Schmidhuber.
\newblock Training very deep networks.
\newblock In {\em Advances in neural information processing systems}, pages
  2377--2385, 2015.

\bibitem[\protect\citeauthoryear{Vinyals \bgroup \em et al.\egroup
  }{2016}]{vinyals2015order}
Oriol Vinyals, Samy Bengio, and Manjunath Kudlur.
\newblock Order matters: Sequence to sequence for sets.
\newblock {\em ICLR}, 2016.

\bibitem[\protect\citeauthoryear{Wang \bgroup \em et al.\egroup
  }{2016}]{wang2016revisiting}
Xinggang Wang, Yongluan Yan, Peng Tang, Xiang Bai, and Wenyu Liu.
\newblock Revisiting multiple instance neural networks.
\newblock {\em arXiv preprint arXiv:1610.02501}, 2016.

\bibitem[\protect\citeauthoryear{Wei \bgroup \em et al.\egroup
  }{2014}]{wei2014scalable}
Xiu-Shen Wei, Jianxin Wu, and Zhi-Hua Zhou.
\newblock Scalable multi-instance learning.
\newblock In {\em Data Mining (ICDM), 2014 IEEE International Conference on},
  pages 1037--1042. IEEE, 2014.

\bibitem[\protect\citeauthoryear{Zhang and Zhou}{2007}]{zhang2007ml}
Min-Ling Zhang and Zhi-Hua Zhou.
\newblock {ML-KNN: A lazy learning approach to multi-label learning}.
\newblock {\em Pattern recognition}, 40(7):2038--2048, 2007.

\bibitem[\protect\citeauthoryear{Zhang and Zhou}{2014}]{zhang2014review}
Min-Ling Zhang and Zhi-Hua Zhou.
\newblock A review on multi-label learning algorithms.
\newblock {\em IEEE transactions on knowledge and data engineering},
  26(8):1819--1837, 2014.

\end{thebibliography}

\end{document}